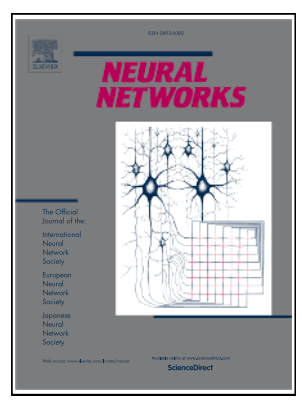

Full Length Article

# Structure from rank: Rank-order coding as a bridge from sequence to structure

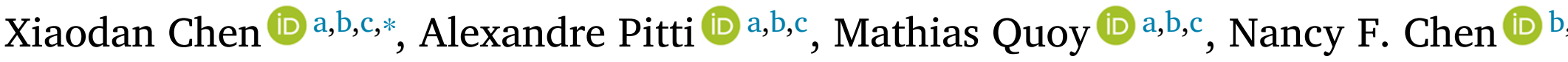
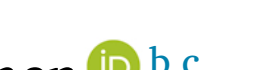
Xiaodan Chen [a,b,c,*], Alexandre Pitti [a,b,c], Mathias Quoy [a,b,c], Nancy F. Chen [b,c]

[a] ETIS, CY Cergy-Paris Université - ENSEA – CNRS, UMR 8051, 2 Av. Adolphe Chauvin, Pontoise, 95300, Ile de France, France
[b] A*Star, 1 Fusionopolis Way, #20-10 Connexis North Tower, Singapore, 138632, Singapore
[c] IPAL (International Research Laboratory on Artificial Intelligence), 1 Fusionopolis Way, #21-01 Connexis (South Tower), Singapore, 138632, Singapore

ARTICLE INFO

*Keywords:*
Rank-order coding
Structure processing
Abstract chunking
Hierarchical structure

ABSTRACT

Understanding how structured sequence information can be represented and generalized in neural systems is key to modeling the transition from acoustic input to emergent structure. In this study, we propose a rank-order based neural network inspired by the STG-LIFG-PMC pathway, modeling the bottom-up transition from acoustic input to abstract rank representation and the top-down generation from that representation to motor execution. Building on previous work in rank coding, we first demonstrate that this model efficiently compresses input while retaining the capacity to reconstruct full utterances from partial cues, revealing emergent structure-sensitive generation process that reflects context-general representations of sensorimotor states, which are later shaped into context-specific motor plans during speech planning. We then show that the network exhibits global-level novelty detection similar to the P3B novelty wave, replicating the global-sequence-sensitive mechanism. As a supplement, we also compare the model's behavior under local (index-level) and global (rank-level) perturbations, revealing robustness to superficial variation and sensitivity to abstract structural violation, key features associated with hierarchical generalization. These results suggest that rank-order coding not only serves as a compact encoding scheme but also captures hierarchical structure in acoustic sequences.

## 1. Introduction

During early vocalization, infants rely on a sensorimotor loop (Choi et al., 2023; Hickok et al., 2011) that tightly couples speech perception and production (Westermann & Reck Miranda, 2004), enabling them to explore the mapping between articulatory gestures and the resulting auditory signals.

Through vocal imitation (Kuhl & Meltzoff, 1996; Wedel & Fatkullin, 2017), or babbling while attempting to imitate sounds within the linguistic environment (Keren-Portnoy et al., 2010; Kröger et al., 2022; Lieberman, 1980; Oller, 1980; Oller & Eilers, 1988; Westermann & Reck Miranda, 2004), infants gradually refine this exploratory mapping into stable internal representations: a foundational inventory of phonetic units (Kuhl & Meltzoff, 1996) (also referred to as prior phonological knowledge Baddeley et al., 1998 or early phonological representations Munson et al., 2011) and a corresponding map for articulatory control (Werker & Tees, 1984) (or vocalic states Kröger et al., 2022). This acquired sensorimotor mapping provides the prerequisite basis for learning combinatorial structures (de Boysson-Bardies et al., 1989; Huybregts, 2017; Kröger et al., 2022; Zuidema & de Boer, 2018), or more specifically, phonological structure for proto-syllabic combinations (Kröger et al., 2022) or proto-phone combinations (Oller et al., 2021) that form protowords (Arbib, 2008; Michon & Aboitiz, 2025), which we term protogrammar (Arbib, 2008, 2012), thereby enabling infants to govern how sounds are sequenced.

However, the representations under which sound sequences are stored and processed remain largely unknown. According to Saffran and Wilson (2003), before regularities can be extracted from auditory input, the sequences must first be segmented and processed. In line with this, Bekinschtein et al. (2009), Dehaene et al. (2015) proposed that the speech stream is initially parsed into discrete chunks. Experimental evidence supports this segmentation process: Saffran et al. (1996, 1999) demonstrated that 8-month-old infants could associate sound chunks with objects after being exposed to just 2 min of fluent speech without pauses. The infants successfully segmented speech (Saffran et al., 1996) or tone (Saffran et al., 1999) stream into sound chunks through statistical learning (Kleinschmidt & Florian Jaeger, 2015; Maye et al., 2002; Schneider et al., 2022), and mapped these segmented sound (word-like) sequences to objects-forming representation with emergent linguistic status. Notably, Estes et al. (2007), Saffran et al. (1996) highlighted that only sequences with an internal structure could be processed as "good words." This internal structure, referred to as "algebraic patterns" (Dehaene et al., 2015) or "abstract algebraic rules" (Marcus et al., 1999), represents a higher level of sequence abstraction by removing specific identities. Such structures, known in human languages as "phonological structures," enable generalization across instances. Dehaene et al. (2015) explored infants' competence in this area by repeating identical sound chunks like *AAAAB* (or *BBBBA*), resulting in the disappearance of P3B, an event-related potential component, associated with the detection of unexpected stimuli. Interestingly, P3B reappeared when a monotonic sequence, such as *AAAAA* (or *BBBBB*), was presented. Similar global-local experiment results were observed in Bekinschtein et al. (2009), El Karoui et al. (2014). This ability to extract abstract rules is crucial for generalizing novel (or global) instances (Hadley, 2009) and is foundational for later generative linguistic capabilities in infancy (Bernal et al., 2009).

* Corresponding author.
*E-mail addresses:* xiaodan.chen@cyu.fr (X. Chen), alexandre.pitti@cyu.fr (A. Pitti), mathias.quoy@cyu.fr (M. Quoy), nfychen@i2r.a-star.edu.sg (N.F. Chen).

However, algebraic patterns represent only one type of pattern. A more meaningful approach may involve replacing algebraic structures with representations that incorporate richer information than simple repetition across elements. One compelling example is the integration of relative temporal information into pattern representation. For instance, an algebraic pattern like *ABA* can be transformed into an ordinal pattern, *121*, which includes timing information, indicating that pattern #1 (or *A*) occurs before pattern #2 (or *B*). Timing information is valuable, as many natural events follow fixed and predictable timing patterns (Dehaene et al., 2015). Moreover, representing temporal patterns is a core function of the brain, with temporal information essential for various types of learning, behavior, and sensorimotor processing (Paton & Buonomano, 2018). Berdyyeva and Olson (2010) demonstrated through experiments that rank-order-sensitive neurons, which fire most strongly to a specific item or action only when it appears in a particular rank within a sequence, are present in several areas of the macaque frontal cortex. During serial action tasks (e.g., making saccades in a fixed sequence of directions) and serial object tasks (e.g., making saccades to a fixed sequence of objects), the majority of these neurons showed preference on the same rank across both tasks. This cross-domain activity suggests that the brain might use a common, abstract coding scheme for representing order, regardless of the specific item.

Research on abstract numerical codes has demonstrated their effectiveness. Botvinick and Watanabe (2007), Botvinick and Plaut (2006) proposed a gain-field network model to simulate cortical working memory, with two input layers: one representing items (e.g., shapes, locations, or verbal items) and the other representing serial order (e.g., rank). The primary feature of the gain-field network is its ability to combine these two inputs multiplicatively (Andersen et al., 1985), thereby integrating item and order information. Building on this approach, Pitti et al. (2020) applied both ordinal coding and the gain-field network to retrieve missing indices within ordered items. Both studies successfully replicated neurons sensitive to ranks. Furthermore, Lebioda et al. (2024) conceptualized the extracted ordinal patterns as grammatical rules governing sequences, which could then be used to generate new sequences.

In line with that, we propose a rank-order-based neural network developed upon the works of Chen et al. (2024), Lebioda et al. (2024), Pitti et al. (2020), modeling the STG-LIFG-PMC pathway (Friederici, 2011; Vigneau et al., 2006), reflecting the bottom-up transition from acoustic input to abstract rank representation and back to motor execution through top-down generation. This work directly extends our previous model (Chen et al., 2024) of the STG-PMC pathway for sensorimotor mapping in early infant vocal imitation (Kuhl & Meltzoff, 1996; Wedel & Fatkullin, 2017), demonstrating how such sensorimotor mapping provides the essential substrate for acquiring structural rules, consistent with developmental evidence that sensorimotor foundations precede and enable linguistic structure learning (Westermann & Reck Miranda, 2004). We test the hypothesis that rank-order coding supports the representation of hierarchical abstract structure. This structure enables a system to encode temporal sequences not as linear strings, but as nested chunks and patterns (Dehaene et al., 2015; Uddén et al., 2019), where relationships are governed by the relative order (rank) of elements. In our context, this structure can be interpreted as a protogrammar-an early, structural form that transforms acoustic patterns into ordered, rule-like configurations without assuming a full linguistic grammar. We designed a series of experiments to target different aspects of this representational capacity. To begin with, we explore the computational efficiency of rank-order coding as a compressed representation, which justifies our choice of abstract chunking length for the following experiments. This computationally derived choice is also consistent with the proposed capacity limits of working memory (Ding, 2025). We then investigate the neural network's capacity to generate 'motor speech sequences' (Hickok & Poeppel, 2007) from reduced or abstract input (e.g., ranks or partial sequential cues), to examine how rank-order coding supports abstract, context-general representations of sensorimotor states during speech planning (Khanna et al., 2024), mimicking the role of the LIFG (Broca's area) (Koechlin & Jubault, 2006). As mentioned in Dehaene et al. (2015), natural events often follow fixed and predictable patterns, which leads us to question whether abstract numerical representation in our neural network exhibits characteristics similar to intrinsic phonological structures or grammar in language. As discussed in Uddén et al. (2019), hierarchical generalization is regarded as convincing evidence for the presence of hierarchical structure or grammar. Therefore, we probe our neural network's potential cognitive relevance by attempting to replicate the global novelty response, analogous to the P3b signal in Dehaene et al. (2015) using pseudo 6-gram stimuli, where such "global response" (Dehaene et al., 2015) implies rejecting sequences violating the grammar (Uddén et al., 2019). As a supplement, we also test the network's sensitivity to deviations at the index (item or local Dehaene et al., 2015) level (or sequential structure Uddén et al., 2019) and its generalization (robustness) at the rank or global (Dehaene et al., 2015) level (hierarchical structure Uddén et al., 2019), to evaluate whether rank-order coding exhibits structural properties consistent with hierarchical organization that may relate to protogrammar processing.

## 2. Methods

### 2.1. Theoretical framework

#### 2.1.1. Neurodevelopmental hierarchies in language processing

Language development progresses from early processing of basic sounds, such as sensorimotor primitives, phonemes, to later stages involving comprehension and production of complex sentences. This developmental sequence mirrors the functional organization of brain regions involved in language processing. Uddén and Bahlmann (2012) describes a rostro-caudal (or anterior-posterior Gough et al., 2005 or anterior/ventral part - posterior/dorsal Vigneau et al., 2006) gradient of abstraction in the left inferior frontal gyrus (LIFG), corresponding to levels of linguistic processing (Hagoort, 2005). Supporting this, Badre and D'Esposito (2009) proposed that neurons in progressively rostral areas of the frontal cortex are specialized for handling more abstract representations and governing more complex rules. This hierarchy is believed to relate closely to the increasing complexity of phonological, syntactic, and semantic stimuli (Vigneau et al., 2006), paralleling the developmental trajectory of language in children (Oller et al., 1976; Repp, 1984; Stark, 1980).

To further explore this hierarchy, the terms of "simple chunks" and "superordinate chunks" (Ding, 2025; Koechlin & Jubault, 2006; Rouault & Koechlin, 2018) are introduced. A superordinate chunk consists of multiple chunks rather than individual sensory events (Ding, 2025), offering a cognitive mechanism for structuring nested information. Rouault and Koechlin (2018) proposed that Broca's area, encompassing BA 44 and BA 45, supports two nested levels of abstract chunking. This two-level chunking structure is considered sufficient for generating a nested tree structure-a fundamental characteristic of language.

(a) Hierarchical organization of rank-order coding

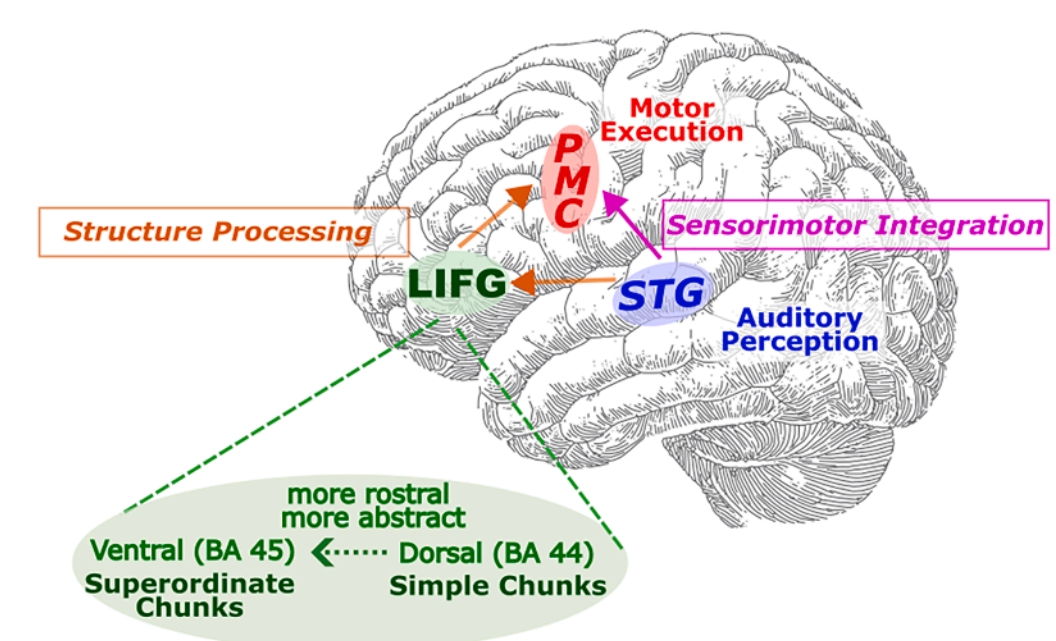

(b) The dual pathway

**Fig. 1.** (a) This illustrates the rank-order framework for transforming waveforms into an abstract representation. Acoustics primitives are categorized, assigned indices by phonetic similarity (e.g., $/AH0/$, $/OW2/$, $/EY1/$ for vowels; $/T/$, $/M/$ for consonants). The indexed elements are then grouped into chunks of three and which are then transformed into rank-order chunks by preserving only their relative ordering. These rank chunks are assigned indices by preserving relative temporal ordering. Finally, 5 rank chunks are integrated into one superordinate chunk, forming a hierarchical structure and yielding a context-general representation. (b) The corresponding dual pathway model, where brain regions are marked with colored borders and pathway activity is color-coded. The pink pathway (STG → PMC) represents the first stream for sensorimotor integration that includes direct sound-to-motor mapping. The orange pathway (STG → LIFG → PMC) performs a **bottom-up** transition (STG → LIFG) from acoustic input *(context-specific)* to an abstract *(context-general)* rank representation (also shown by left orange arrow in (a), and a **top-down** projection (LIFG → PMC) from this rank chunk to a concrete motor sequence *(context-specific)* for articulation (also shown by right orange arrow in (a).

For better comprehension, we will further explore this hierarchical organization through an example based on rank-order coding.

Fig. 1a illustrates the process of deriving rank-order chunks, which are later transformed into a superordinate chunk from utterances input. The phonemes are based on the ARPA phone set (McAuliffe & Sonderegger, 2024) and are visualized using Praat (Boersma & Weenink, 2001). To begin with, phoneme indices are attributed based on phonetic similarity: vowel sounds are placed closer together than consonants. This organization is predicated on the observation that neurons located close to each other tend to encode similar phonetic properties, forming local computational units for utterances processing (Leonard et al., 2023; Meisenhelter & Rutishauser, 2022; Mesgarani et al., 2014; Paulk et al., 2022). Consequently, each phoneme (or its corresponding index) serves as a basic unit of information that can be processed and stored. These indexed phonemes are first grouped into sequences of length 3, referred to as simple chunks. They are then transformed into rank chunks, which capture both relative and sequential information based on phonetic characteristics. To handle tied elements within index sequences, we resolve ties by assigning different ranks based on the order of appearance, ensuring that the temporal structure is maintained. These rank chunks are assigned indices by preserving relative temporal ordering. Finally, these rank chunks are combined into a superordinate chunk, comprising 5 elements in our example. This superordinate chunk represents a higher-level unit of meaning or organization, encapsulating more information than individual rank chunks. This hierarchical organization illustrates how a 38-s spoken word sequence is progressively structured, evolving from simple units (phonemes) to complex formations (superordinate chunks) within our rank-order processing framework. In the next section, we shift from this illustrative, phoneme-based example to the pre-phonological neural processes that form our model's foundation. While the figure uses phonemes for clarity, our approach models acoustic input in the form of MFCCs as basic units, capturing early auditory-motor patterns from which rank-order chunks consistent with phonological structure can emerge.

### 2.1.2. Dual pathways for language processing

To further explore this hierarchical organization of speech processing, Fig. 1b illustrates the 'two pathways' model inspired by Friederici (2011), Hickok and Poeppel (2007), Rouault and Koechlin (2018). The STG, depicted in blue, is located in the temporal lobe and is responsible for auditory processing (Friederici, 2011). The PMC, shown in red, is situated in the frontal lobe and plays a key role in speech planning and speech articulation (Hickok & Poeppel, 2007). The LIFG (especially Broca's area) is crucial for sequencing sounds and building hierarchical or ordered structures (Hagoort, 2005).

The first pathway (illustrated in pink) connects the auditory processing regions of the STG to the Premotor Cortex (PMC). Neurobiologically, this circuit is the core substrate for sensorimotor integration in speech, facilitating sound-to-motor mapping (Vigneau et al., 2006). The second pathway, illustrated in orange, links the temporal cortex to Brodmann Areas 44 (BA 44) and 45 (BA 45) in the left inferior frontal gyrus (LIFG), supporting more complex language processes, such as phonological working memory storage (Paulesu et al., 1993) and structure-function mapping (Tyler & Marslen-Wilson, 2007). BA 44 is particularly associated with simpler chunk processing, while BA 45 is involved in the integration of superordinate chunks, reflecting increasingly abstract representations within the rostro-caudal gradient of abstraction as discussed previously.

Within this framework, our model implements a complete cycle of speech processing, modeling the bottom-up transition from acoustic input to abstract representation and the top-down projection from that to motor production.

Specifically, the first pathway performs essential sensorimotor integration at the pre-phonological level, thereby generating the internal state (Breault et al., 2023) that enable higher-order processing in the second pathway. Sound sequences are processed in the STG and transformed in the PMC into internal state, namely index chunks that encode the sensorimotor identity of the sound, consistent with findings that sensorimotor regions encode internal states rather than direct motor commands (Breault et al., 2023). These index chunks are then transformed into rank representations within the LIFG, reflecting a bottom-up transition from acoustic input to a context-general representation. This rank code, embodying protogrammar, is then operated on to generate a complete motor plan. This plan, in the form of a structured index sequence, is realized for articulation in the Primary Motor Cortex (M1), completing the top-down projection to motor production.

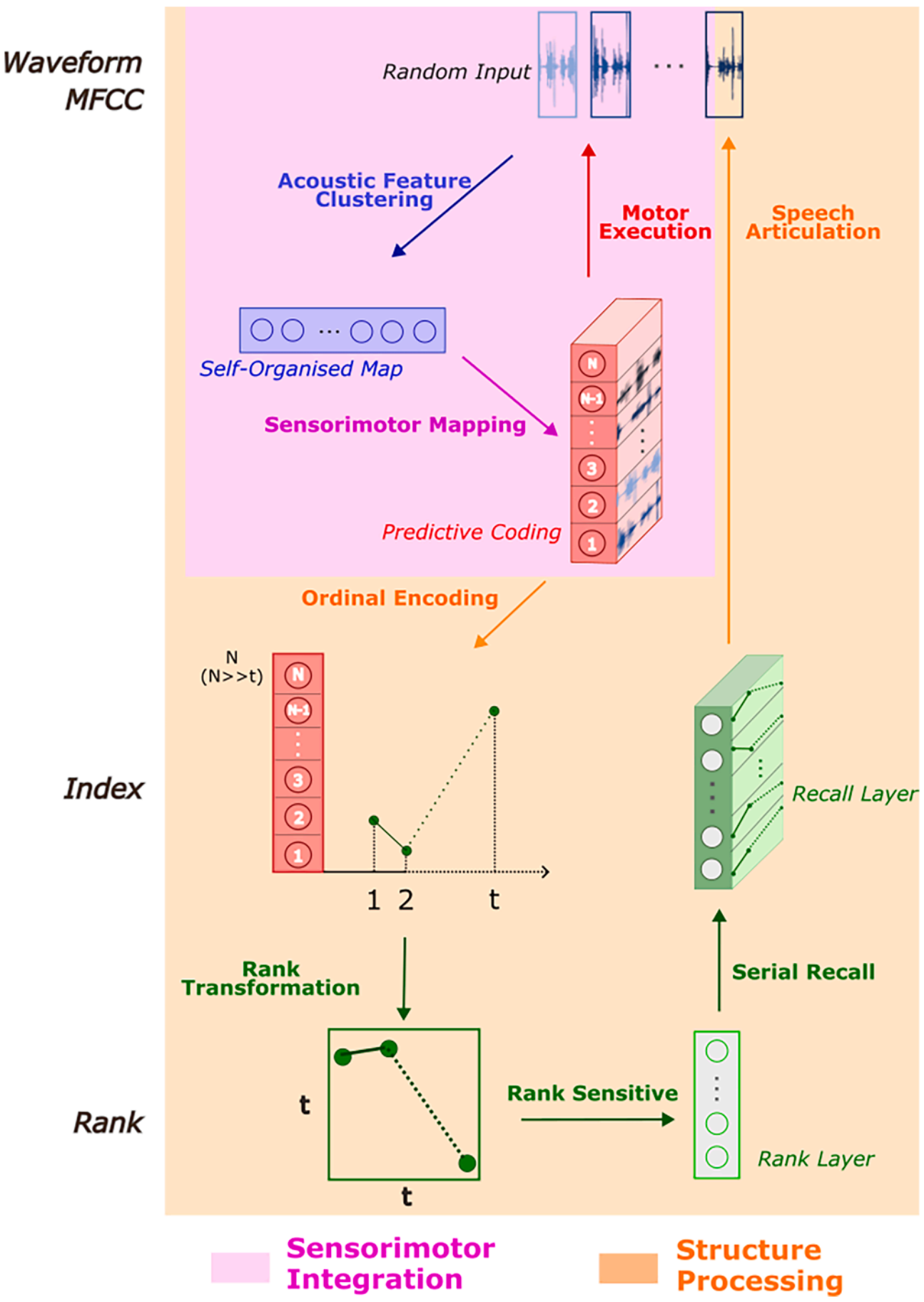

**Fig. 2.** Brain-inspired neural network diagram. The architecture consists of two primary loops, color-coded to match the pathways in Fig. 1. Pink Pathway (Sensorimotor): Models the rapid auditory-motor loop from the Superior Temporal Gyrus (STG) to the Premotor Cortex (PMC), enabling elementary level sensorimotor mapping. Orange Pathway (Hierarchical Processing): Supports higher-level processing, where acoustic inputs are transformed into sensorimotor state index chunks. These are converted into abstract rank codes in the Left Inferior Frontal Gyrus (LIFG) and then recalled into complete, structured motor plans, which tune the PMC output. The model processes inputs through three representational levels in Fig. 1a: acoustic (MFCC), index, and rank.

In the next section, we will discuss the proposed computational models of these two pathways.

## 2.2. Model implementation

In this section, the diagram illustrating the key components of the system will be presented first, providing a visual overview of the network structure. This will be followed by a detailed discussion of the architecture, which delves into the specifics of layer configurations, connections, and operational mechanics.

### 2.2.1. Neural network diagram

Building on the previous discussion, we propose a neural network model with two components that correspond to the dual pathways outlined earlier, with each color representing a specific pathway in Fig. 1b. On the left side of the figure, three levels of representation are indicated: waveform or Mel Frequency Cepstral Coefficient (MFCC), index, and rank, aligned with the hierarchy presented in Fig. 1a.

The first loop, marked by pink borders, models the temporal-frontal auditory-motor network extending from the Superior Temporal Gyrus (STG) to the Premotor Cortex (PMC). This loop corresponds to the 'sensorimotor mapping' pathway shown in both Fig. 1a and b.

The input layer receives MFCCs, which are passed to a Self-Organizing Map (Kohonen, 1982, 1998) (SOM) that categorizes the input, resulting in a corresponding winner neuron for each frame. This winner neuron, along with the SOM output, activates a single neuron in the subsequent layer. Through predictive coding, the first loop enables the rapid elementary sensorimotor integration. This part is based on the neural network described in Chen et al. (2024). For comprehensive details, we refer the reader to that work.

The second loop, highlighted in orange, supports higher-level structure processing. It modulates and tunes motor cortex outputs via connections to Broca's area, corresponding to the hierarchical processing pathway illustrated in Fig. 2. In our model, this pathway is instantiated by the projection from the STG ($Y_{som}$) to the PMC ($Y_{pred}$), which transforms auditory categories into internal state (Breault et al., 2023), namely simple index chunks that constitute potential motor plans as shown in Fig. 1a. Here, hierarchical processing transforms the input into an abstract rank representation. This representation does not generate the motor plan itself, but rather tunes the ongoing planning process in the Premotor Cortex. This tuning occurs via the iterative loop where the internal state in the PMC ($Y_{pred}$) is refined by the LIFG ($Y_{rank}$, $Y_{recall}$), resulting in a complete, grammatically-structured motor plan.

Let **t** denote the number of input time steps, and **N** the number of neurons in the predictive coding layer. During ordinal encoding, the model chunks the indices of the most strongly activated neurons ("winner-take-all" units) at each time step, forming an index chunk of dimension **t**, where each index lies in the range $[0, N-1]$. This simple index chunk, while compact, is sensitive to the absolute identity of the neurons. To extract a more robust, relational code, the model applies a rank-order transform. This involves computing the relative ordering of the indices within the chunk, resulting in a rank chunk of the same dimension **t**. This transformation abstracts away from specific neuron identities to capture the invariant ordinal structure of the activation pattern, forming a context-general representation suitable for hierarchical processing. The resulting rank-encoded representation is then passed through a rank-sensitive transformation layer, implemented as a fixed or learned modulation function. The activation of this rank-sensitive layer projects to a downstream content-addressable memory-like layer (Hopfield, 1982), which recall (Wickelgren, 1967) the most closely matching stored prototype, effectively decoding the rank pattern into a learned index sequence. The rank and recall layers together form a pattern completion system that can reconstruct stored serial sequences from noisy or partial rank patterns, analogous to associative memory systems such as Hopfield networks (Hopfield, 1982) or serial position-to-item associations in short-term memory (Wickelgren, 1967). The recalled serial chunk is then used to activate specific motor command neurons, resulting in the generation of an output signal such as an MFCC sequence.

Again, the first loop is designed to mimic infant perception: the initial categorization of auditory input activates corresponding sensorimotor representations, a process that precedes and enables the subsequent chunking of the signal into structured units. As a result, the second pathway develops later than the first as mentioned earlier, so during the training of the second component, the first component remains frozen. Training details will be presented in the next section, along with the detailed neural network architecture and mathematical representations.

### 2.2.2. Neural network architecture

Following the previous section, Fig. 3 illustrates the model's learning process for a 16-min sound stream. The input is first converted into 41,739 frames of 20-dimensional MFCCs and processed through the sensorimotor pathway, a necessary precursor to the chunking stage as discussed earlier. As its role is to provide a stable foundation for this process, the sensorimotor pathway's weights remain frozen during the training of the second pathway. The number of neurons in this pathway, 2000 for $Y_{som}$ and 5256 for $Y_{pred}$, is indicated in Fig. 3. This sensorimotor pathway is based on the neural network described in Chen et al. (2024), to which we refer the reader for detailed information. Consequently, this section focuses solely on the hierarchical processing pathway.

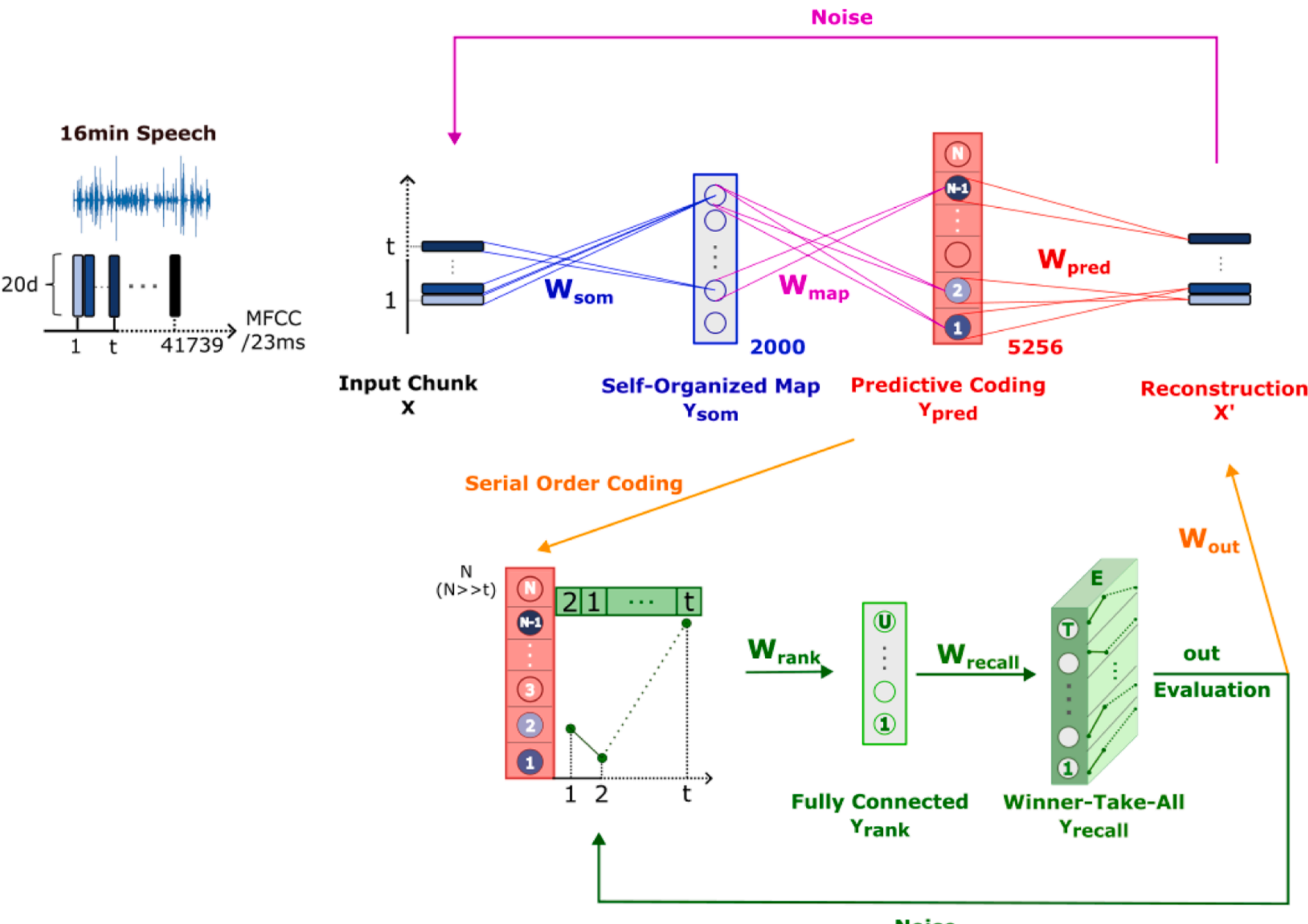

**Fig. 3.** Neural network architecture.

As shown in Fig. 3, let $t$ represent the chunk length (or time step), $T$ the number of input chunks, $U$ the number of unique rank patterns. Let the input sequence be $X = \{x_1, x_2, \dots, x_T\}$, where each $x_i \in \mathbb{R}^t$ represents one chunk of acoustic features with $i \in \{1, 2, \dots, T\}$ indexing training examples. Each input chunk $x_i$ is processed sequentially through the $Y_{\text{som}}$ and $Y_{\text{pred}}$ layers. Through serial order coding (Gordon & Meyer, 1987), we chunk the $Y_{\text{pred}}$ output into a discrete index sequence by applying winner-take-all selection across the neuron activations at each time step:

$$I_i = \left\{ \arg\max(Y_{\text{pred}}^{(i,1)}), \arg\max(Y_{\text{pred}}^{(i,2)}), \dots, \arg\max(Y_{\text{pred}}^{(i,t)}) \right\} \tag{1}$$

where $I_i \in \mathbb{N}^t$ is the index chunk for the $i$th input example with $k \in \{1, 2, \dots, t\}$ indexing positions within each chunk. Each $\arg\max(Y_{\text{pred}}^{(i,k)}) \in \{1, 2, \dots, N\}$ represents the winning neuron index at position $k$, where $N$ is the total number of neurons in the prediction layer. The index sequence $I_i$ is then transformed into a rank representation before entering the $Y_{rank}$ layer. For each $I_i$, we compute its rank sequence $R_i = \{r_{i,1}, r_{i,2}, \dots, r_{i,t}\}$, where $r_{i,k}$ represents the rank of the $k$th element within $I_i$ when sorted in ascending order. Formally:

$$R_i = \text{argsort}(\text{argsort}(I_i)) \tag{2}$$

This operation is efficiently implemented using NumPy's argsort() function twice. The resulting $R_i \in \mathbb{N}^t$ contains integer ranks in the range $[0, t-1]$. Please note that while our implementation uses 0-based ranks [0, t-1] for computational efficiency, all visualizations and conceptual explanations use 1-based ranks $[1, t]$ for cognitive clarity and alignment with conventional ranking systems. We then extract the unique rank patterns across all training examples, defining $R_{\text{unique}} = \{\mathbf{r}_1, \mathbf{r}_2, \dots, \mathbf{r}_U\}$, where $U \le T$ represents the number of unique rank patterns observed during training. Each $\mathbf{r}_j \in \mathbb{N}^t$ is a distinct rank sequence that defines the receptive field of the $j$th neuron in the rank layer, with $j \in \{1, 2, \dots, U\}$ indexing both the unique patterns and the corresponding rank-sensitive neurons. The rank layer is a fully-connected layer with weights $W_{\text{rank}} \in \mathbb{R}^{U \times t}$, updated according to:

$$W_{\text{rank}}[j,k] = \frac{1}{\mathbf{r}_j[k] + 1} \tag{3}$$

The recall layer weights $W_{\text{recall}} \in \mathbb{R}^{T \times U}$ are computed as:

$$W_{\text{recall}}[i,j] = \sum_{k=1}^{t} W_{\text{rank}}[j,k] \cdot \frac{1}{R_i[k] + 1} \tag{4}$$

In this formulation, each row in $W_{\text{rank}}$ encodes a unique, modulated rank pattern (or internal representative Wickelgren, 1967) observed during training, while $W_{\text{recall}}$ stores the projection of each training input into the rank-sensitive space. The rank layer activation is computed using matrix multiplication as:

$$Y_{\text{rank}} = \frac{1}{R+1} \cdot W_{\text{rank}}^{\top} \tag{5}$$

where $R \in \mathbb{N}^{T \times t}$ thus $Y_{\text{rank}} \in \mathbb{R}^{T \times U}$. The rank layer is sensitive to rank patterns, as its neurons activate most strongly when the modulated input rank pattern aligns with the learned patterns. The recall layer computes a similarity-based response between the input projection $Y_{\text{rank}} \in \mathbb{R}^{T \times U}$ and each stored pattern (or internal representative Wickelgren, 1967):

$$Y_{\text{recall}} = \frac{1}{1 + \text{PairwiseDist}(W_{\text{recall}}, Y_{\text{rank}})} \tag{6}$$

where PairwiseDist$(A, B)$ computes the Euclidean distance between all rows of $A$ and all rows of $B$, yielding $Y_{\text{recall}} \in \mathbb{R}^{T \times T}$, an activation pattern where each element reflects the similarity between the current input and the corresponding stored representation. Let $E = \{e_1, \dots, e_T\}$ be the repository of index sequences associated with training data, where each $e_i \in \mathbb{N}^t$ corresponds to $I_i$. The output is determined by:

$$\text{out} = e_{Y^*_{\text{recall}}} \tag{7}$$

where $Y^*_{\text{recall}} = \arg\max(Y_{\text{recall}})$ denotes the index of the winning stored pattern for each input. For motor output reconstruction, the retrieved index sequence out $= \{\text{out}^1, \dots, \text{out}^t\}$ represents a fully specified motor plan. In the motor execution phase, this plan is projected via the output weight matrix:

$$W_{\text{out}} = W_{\text{pred}} \tag{8}$$

$$X' = \left[W_{\text{out}}[\text{out}^k]\right]_{k=1}^{t} \tag{9}$$

producing $X' \in \mathbb{R}^{t\times d}$, where $d$ is the dimensionality of the acoustic features. This process generates coordinated MFCC features, completing the top-down projection from rank representations to motor production. The mechanism is analogous to speech articulation, where the brain executes a motor plan as coordinated articulatory gestures to produce acoustic output.

This whole decoding process models a biologically inspired pathway in which Broca's area plays a central role in mapping abstract representations to articulatory motor plans. The rank-based index code can be viewed as a compressed representation of sequential structure, analogous to hierarchical or syntactic primitives, which are decoded in Broca's area to drive specific motor outputs in the articulatory cortex. The reactivation of motor sequences via indexed memory aligns with views of Broca's area as a hub for speech planning (Friederici, 2011; Hagoort, 2014).

## 3. Experiments

In this part, we first assess the compression efficiency of rank-order representations to reason our choice of abstract chunking length. Secondly, we examine the network's ability to reconstruct utterance sequences from reduced or abstract cues. We then investigate the cognitive plausibility of rank-order coding by attempting to reproduce the "global response" reported in Dehaene et al. (2015). Finally, we evaluate the system's sensitivity to local (item-level) deviations and its generalization to higher rank (global) level structure (hierarchical structure).

### 3.1. Rank as a high-efficiency compression space

To begin with, we explore the efficiency of data compression using rank as a compression method. The utterances dataset we used is in English, derived from the Librispeech corpus (Panayotov et al., 2015). For our analysis, we employ Mel-Frequency Cepstral Coefficients (MFCCs) as input features. Each input is a 20-dimensional vector corresponding to a 23ms speech segment, as shown in Fig. 3, effectively capturing the essential characteristics of the speech sequence in our dataset.

As for the transformation from MFCC chunks to index chunks, the input indices for this study were obtained using the vocal imitation (Kuhl & Meltzoff, 1996) framework, as detailed in our previous work (Chen et al., 2024). Differently however, a significant change in this implementation lies in the design of the $Y_{pred}$ layer (predictive model), which incorporates a topology ensuring that neurons located close to each other in their representational space produce acoustically similar sounds. This is supported by Mesgarani et al. (2014), Paulk et al. (2022), Meisenhelter and Rutishauser (2022), which show that neurons in the auditory and prefrontal cortex topographic arrangement according to articulatory-phonetic similarity.

To achieve this, the neurons in the PMC layer are organized to correspond to those in the STG layer, which is a one-dimensional self-organizing map (SOM). This structural relationship guarantees that the mapping between the two layers respects the input space's inherent topology. The topology of both layers is depicted in Fig. 4. For reference, results from our previous work are shown in Fig. 4a, while Fig. 4 visualizes the organization of the 20-dimensional neurons in the newly pre-trained predictive model, reduced to a two-dimensional space. The accompanying color bar on the left represents the indices of the neurons, revealing that neurons with similar indices share similar colors.

As shown, dots with similar colors are closely clustered in the two-dimensional space, demonstrating that the topological structure has been preserved. This spatial arrangement confirms that the topology within the PMC layer aligns with the organization of the STG layer, providing evidence of effective topological mapping in the model. This setting aligns with Leonard et al. (2023), which shows that neurons close to each other tend to encode similar phonetic properties, forming local computational units for speech processing.

Once the predictive production model is well-trained, we can obtain the corresponding index chunks or rank chunks using the methods illustrated in Figs. 1a, 2, and 3.

Fig. 5a–d depict the growth of dimensions in terms of MFCCs, index chunks, and rank chunks. The horizontal axis represents the duration of the sound dataset in minutes of speech, while the vertical axis shows the logarithmic scale of the counts for MFCCs, index chunks, and rank chunks. As shown, when chunk lengths are below 8, the number of MFCCs and index chunks increases rapidly. In contrast, the rank chunk count remains relatively stable, demonstrating substantial compression from the MFCCs space to the index space and even greater compression to the rank space. Once the chunk length reaches 8, the growth rate of rank chunks begins to converge with that of index chunks, the growth rate of rank chunks starts to converge with that of index chunks . To further isolate the impact of sequence length from dataset size, we present Fig. 5e, which shows that rank chunk counts are more sensitive to sequence length than index chunk counts. Consistent with earlier observations, when the chunk length approaches 10, rank-order representations begin to lose their abstraction advantage and converge with the index space in representational complexity.

For the remaining experiments, we adopt a chunk length of 6, as it offers a good balance between sensitivity to structure and representational efficiency. A shorter chunk length 4 showed minimal improvement with increasing data, indicating potential limitations in capturing higher-order structure. In contrast, a longer chunk length 8 yielded lower compression efficiency, suggesting redundancy. Notably, with a chunk length of 6, performance plateaued around 16 min of data, suggesting the emergence of stable, grammar-like structural generalization. This result aligns with psychological theories of working memory capacity, which posit that a chunk optimally contains fewer than seven sensory events (Ding, 2025).

### 3.2. Continuous sequence generation using a fixed-size sliding window

This experiment investigates our model's capacity for sustained, autoregressive generation from partial cues, testing its ability to reconstruct full sequences through an emergent, structure-sensitive process. The central hypothesis is that this generation is grounded in a stable, higher-level rank chunk, which provides a context-general representation of sensorimotor states. This abstract representation is subsequently shaped into context-specific sequences during speech planning and finally for speech articulation. We evaluate this by challenging the model in a closed-loop inference setting, designed to assess its ability to generate a complete multi-step sequence using only initial cues, relying entirely on its own predictions without access to future ground-truth.

This autoregressive generation employs a sliding-window mechanism to manage the iterative prediction process. Supposing the chunk length is $L$ and the model is initialized with a context of $L - N(N < L)$ starting indices and generates the subsequent block of $N$ indices, with higher level representation rank as a protogrammar that tunes the generation. The window then advances by a stride of $N$, incorporating the newly predicted $N$ indices into the context for the next step. This fixed window autoregressive loop is repeated for $K$ steps. The integrity of this generation process is founded on a data preparation pipeline designed to mirror the inference dynamics. For training dataset, sequences are segmented into overlapping windows of length $L$ with a stride of $N$. These concrete neural network parameters will be detailed after the pilot experiment for the main autoregressive generation task.

The generation process employs a predictive coding-inspired mechanism operating within $K$ iterations per autoregressive step. At each

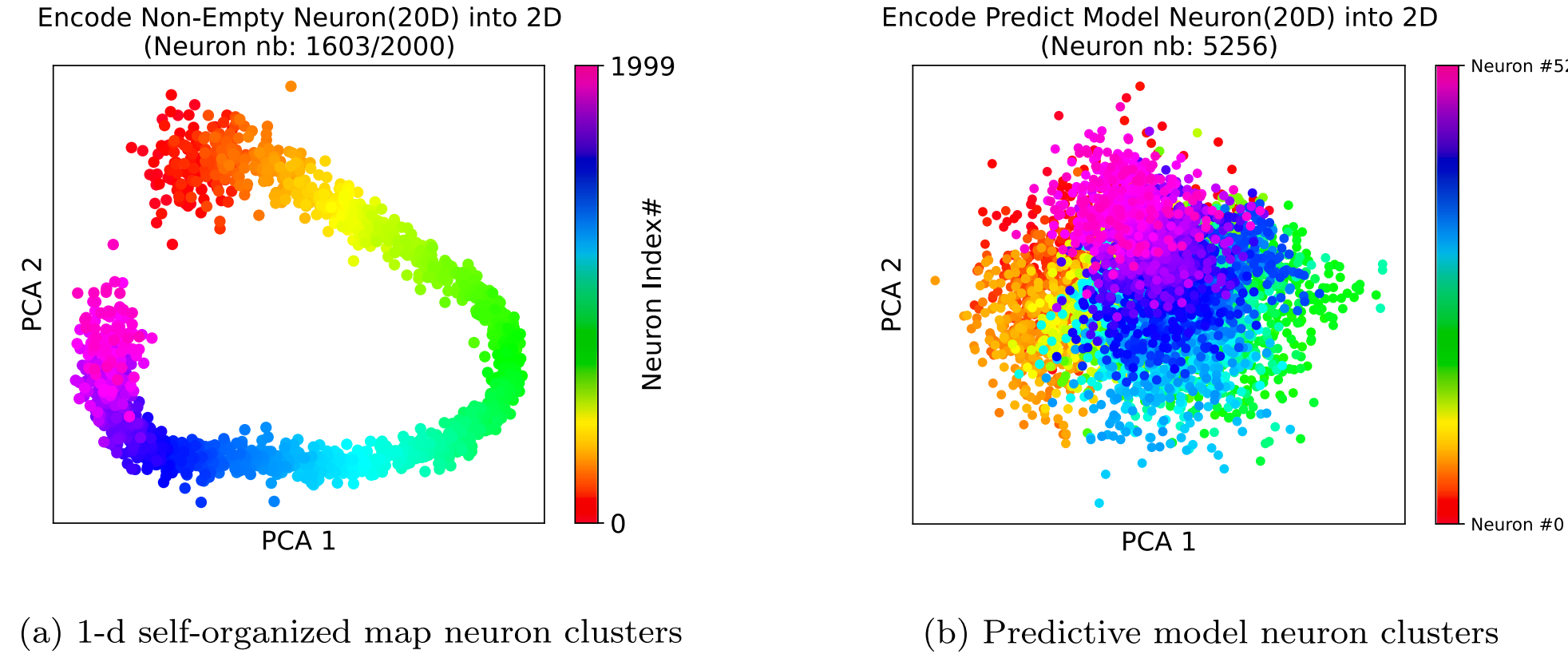

(a) 1-d self-organized map neuron clusters  (b) Predictive model neuron clusters

**Fig. 4.** Neuron clusters in STG ($Y_{som}$) (left) and PMC ($Y_{pred}$) (right) layers. (a) illustrates the topology of both layers, with the 20-dimensional neurons of the new pre-trained predictive model reduced to two dimensions. For comparison, Fig. 4a shows results from our previous work. The color bar represents neuron indices, with similar indices sharing similar colors. Closely clustered dots of similar colors in the two-dimensional space confirm that the topological structure is preserved. This spatial arrangement validates the alignment between the STG and PMC layers, demonstrating effective topological mapping in the model.

(a) chunk length: 4  (b) chunk length: 6

(c) chunk length: 8  (d) chunk length: 10

(e) Dataset size: 16min

**Fig. 5.** (a)-(d): Growth of dimensions in terms of MFCCs, unique index chunks, and unique rank chunks across different dataset sizes. (e): Comparison of MFCCs, unique index chunks, and unique rank chunks across different chunk lengths with a dataset size of 16 min.

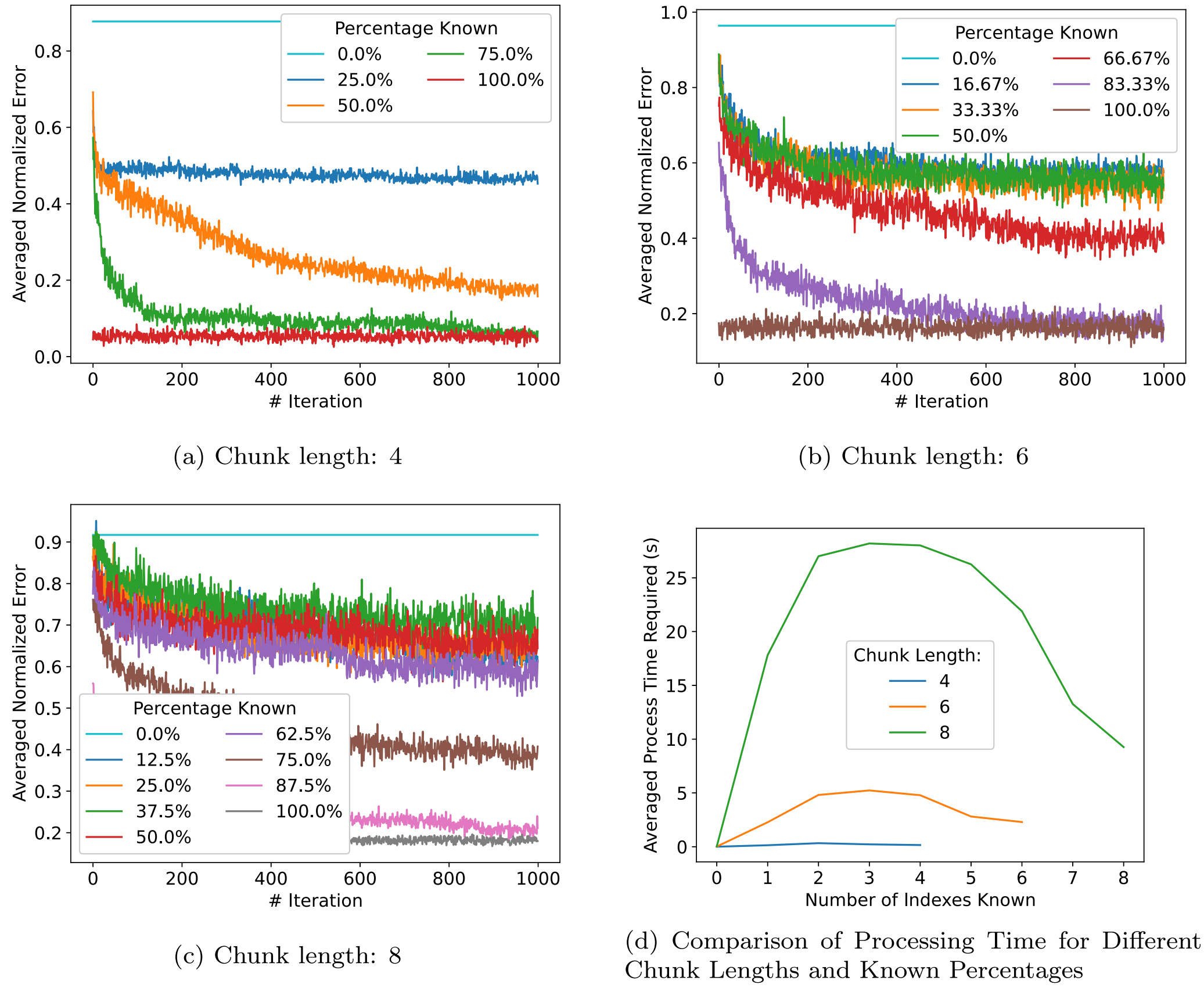

**Fig. 6.** Performance analysis of index reconstruction: (a)-(c) show error convergence across iterations for different chunk lengths, revealing that higher known percentages do not always improve accuracy, particularly for longer sequences. (d) shows the corresponding processing times, demonstrating a non-linear relationship between known indices and computational efficiency.

step, the network receives a context of $L - N$ known indices combined with $N$ random placeholders, which are then projected into the compressed rank-order space via the $Y_{rank}$ layer. The subsequent $Y_{recall}$ layer then generates a candidate output sequence through iterative refinement. Critically, this candidate sequence is accepted only if the candidate sequence, when projected back to $Y_{rank}$ and then $Y_{pred}$, gets a higher activation. Additionally, apart from the convergence limitation of 1000 iterations, this loop stops once the accepted candidate includes the originally provided known indices. The final element of the accepted candidate sequence is then taken as the new predicted index. This process ensures that the model's output converges towards a state of high internal consistency, as defined by its learned rank-based grammar, while operating within a fixed computational budget.

As a pilot experiment, we first assess the effectiveness of reconstructing original index sequences across different percentage of known indexes provided in one step across 1000 iterations. Performance is compared by varying the percentage of known indices within the target sequences and examining different sequence lengths: 4 (Fig. 6a), 6 (b), 8 (c). The percentage of known indices ranges from 0% to 100%, calculated as the number of known indices divided by the total number of indices. Each test involves 100 examples.

As shown in Fig. 6, a higher percentage of known indices does not always correlate positively with improved performance, particularly for longer sequences. For instance, when the sequence length is 6, the performance for scenarios with 1, 2 or 3 known indices is similar, as shown in Fig. 6b. However, regardless of the sequence length, when only one index remains unknown, the reconstruction error consistently converges to zero after 1000 iterations. Additionally, Fig. 6d illustrates the processing time required for each of these scenarios. Similar to the previous results, the findings reveal that having more known indices does not always improve efficiency, as it does not consistently reduce processing time. This observation highlights that the relationship between the number of known indices and processing time is not strictly linear.

Guided by the pilot experiment, we adopted a chunk length $L = 6$ and a step size $N = 1$ for subsequent autoregressive generation task. From the original 41,739 MFCC, this configuration produced 41,734 overlapping training index sequences. To ensure the model learned meaningful transitions, we filtered out sequences with only identical elements, resulting in a final training set of 40,667 index sequence samples. In this setting, a long sequence (e.g., length 10) will be reconstructed incrementally, starting with 5 known indices. This fixed-size sliding window (e.g., length 6) shifts by one index at a time to include the newly generated index. Similar to the pilot experiment, the system retrieves the last index within 1000 iterations. This newly retrieved index is then added to the input for the subsequent retrieval round as a known index.

Results are illustrated in Fig. 7a, b. Target long sequences are displayed as thick gray lines for better visibility when no error is present. Known indices are highlighted in red and marked in green border for clarity, while generated indices are represented by dashed blue dots. The sliding window's incremental process is illustrated with smaller visualizations in Fig. 7a and b, where the light green frames here indicate the current partial cues. As shown in the results, sequences of length 10 were successfully generated from only 5 known indices, requiring five retrieval rounds. Notably, this approach can be extended to generate longer sequences, as demonstrated later.

So far, the model has demonstrated its capability to generate consecutive indices continuously. This rigorously tests the rank-based grammar's ability to maintain structural fidelity and resist the drift typically caused by recursive prediction.

We next evaluate the system's capacity for continuous sound generation, testing the robustness of the rank-based representational scaffold in a complete perception-production pipeline. Using the same dataset employed for index sequence generation, we first fine-tune the pretrained predictive model via continuous learning as desribed in Chen et al. (2024) on the test data, thereby enhancing its performance for sound reconstruction.

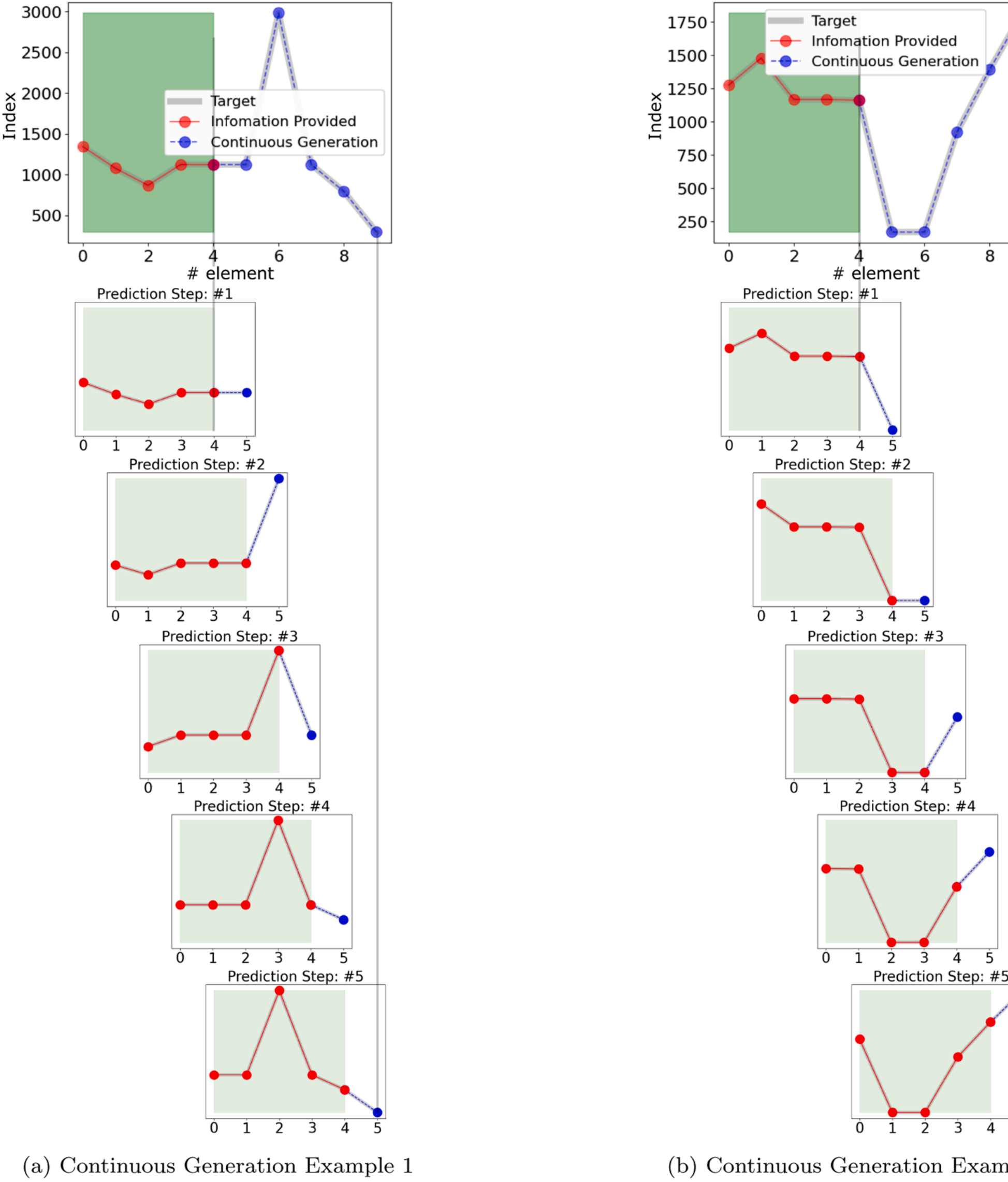

(a) Continuous Generation Example 1 (b) Continuous Generation Example 2

**Fig. 7.** Continuous index generation. Chunk length: 6; Number of index given: 5. In this process, the dark green frames indicate the initially provided indices, while the light green frames represent the sliding window, which moves with a stride of 1 each timestep. The sliding window iteratively retrieves the upcoming index within the current chunk and includes it in the input for the subsequent retrieval round. This iterative process enables the generation of an long sequence from a limited set of initial indices.

The entire process modeling the bottom-up transition from acoustic input to higher level rank chunk and the top-down projection from that to motor production, aiming to test the effectiveness in input compression while retaining the capacity to reconstruct full utterances from partial cues, revealing an emergent, structure-sensitive generation process.

This process is divided into two stages. In the perception stage, sound sequences are processed in short temporal windows (e.g., 23 ms) and categorized in the STG ($Y_{som}$). These sensory patterns are then transformed in PMC ($Y_{pred}$) into partial index chunks, forming an internal state (Breault et al., 2023).

The subsequent production stage unfolds as speech planning and speech production. The internal state (partial index chunk) is concatenated with a random index ($N = 1$) and serves as input to a higher-order planning circuit. This engages the rank-based grammar in the LIFG ($Y_{rank}$), reflecting a bottom-up transition from acoustic input to a context-general representation. This abstract structure is then operated on by the recall layer ($Y_{recall}$) to generate a complete sequence in a autoregressive way, similar to the previous experiment. We interpret this loop, from concrete indices to abstract rank and back to concrete index sequence, as the core speech planning mechanism. Through iterative autoregressive generation, the model shapes the context-general rank patterns into context-specific index sequences for speech planning. Finally, in the motor execution phase, this fully specified index plan is projected via $W_{out}$ to generate MFCC features, completing the top-down projection to speech production. This process is analogous to speech articulation, where the brain executes a motor plan as coordinated articulatory gestures. This integrated framework demonstrates how abstract, rank-based representations bridge perception and action, enabling fluent speech sequence generation from partial acoustic input.

The results of the continuous sound generation experiment are summarized in Fig. 8. In Fig. 8a and b, following the same paradigm as in previous figures, the model successfully reconstructs the complete target index sequences (lengths 19 and 36, respectively) from a minimal

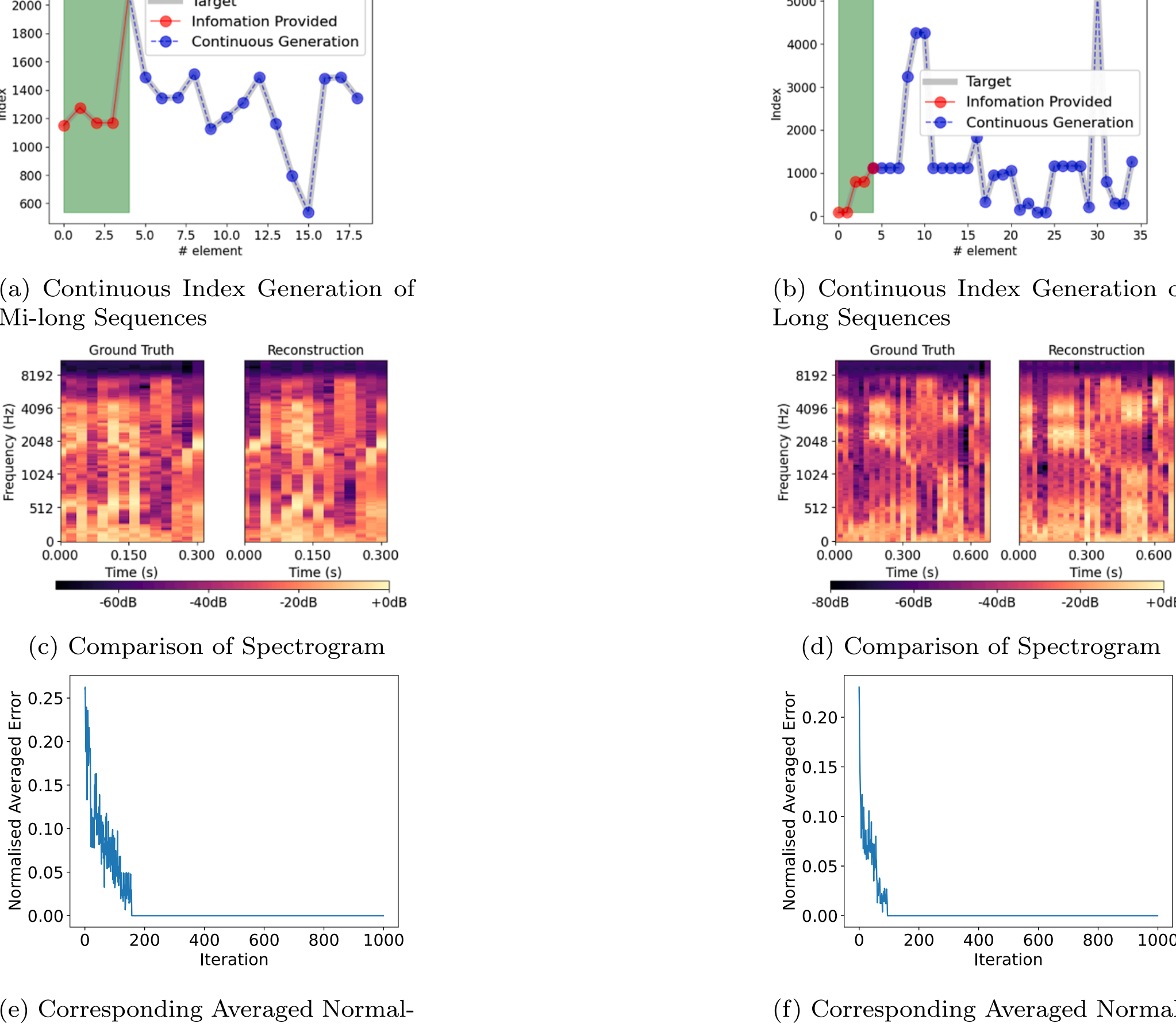

(a) Continuous Index Generation of Mi-long Sequences

(b) Continuous Index Generation of Long Sequences

(c) Comparison of Spectrogram

(d) Comparison of Spectrogram

(e) Corresponding Averaged Normalized Chunk Error Tendency

(f) Corresponding Averaged Normalized Chunk Error Tendency

**Fig. 8.** (a)-(b): Red dots in green frame: the initial input; blue dots: Generated indices; blue dash line: generated index pattern ; Gray lines: target index pattern. The model successfully reconstructs the full index sequences (lengths 19 and 36, respectively) from only 5 initial indices derived from the input MFCCs. (c)-(d): Spectrograms derived from the reconstructed MFCCs. The reconstructed spectrograms preserve the main time-frequency structure and key spectral events of the ground truth. The primary differences lie in smoother textures and reduced high-frequency detail, showing that while the global structure is well reproduced, some fine-grained spectral information is lost. (e)-(f): Reconstruction error over iterations. The error converges to zero by 200 iterations, demonstrating an effective and robust generation process.

initial context of only five indices, which are derived from the input MFCCs (marked by dark green frames). To evaluate the model's capacity for generating longer sequences, the target length is increased from the 10 indices used previously to 19 and 36 indices in this experiment. A perfect alignment is observed between the generated indices (blue dots, connected by dashed lines) and the target chunks (thick gray lines), indicating a reconstruction error of zero. This accurate index generation translates to high-fidelity sound reconstruction, as evidenced by the spectrograms in Fig. 8c and d. In both examples, the reconstructed spectrograms closely follow the ground-truth patterns, preserving the main time-frequency structure, major spectral transitions, and overall energy distribution. This shows that the model reliably captures the essential acoustic content and maintains good temporal alignment. The main differences appear in the finer details: the reconstructions are slightly smoother, with less pronounced high-frequency variations and a mildly compressed dynamic range. Overall, the results indicate that the model produces spectrograms that are structurally accurate and acoustically consistent, while still missing some of the sharper local spectral features present in natural speech. All corresponding audio samples are available at Chen et al. (2025). Finally, the robustness of this process is quantified in Fig. 8e and f, which plot the reconstruction error over iterations. The error converges to zero stably and consistently, typically within 200 iterations, demonstrating the effectiveness and reliability of the autoregressive generation loop. As shown, despite this significant increase in sequence length, our proposed neural network demonstrates its capability to reconstruct the sequences effectively.

### *3.3. Global-sequence-sensitive mechanism: Detection of rank-level violation*

To investigate whether our neural network exhibits global-level sensitivity to novel sequences, we draw inspiration from the P3b novelty signal described in Dehaene et al. (2015). In that study, the reappearance of the P3b component was observed in response to a globally novel sequence, suggesting sensitivity to higher-order regularities. In this section, we aim to reproduce this "global response" using our model.

Fig. 9 presents the waveform of a sound stream segmented into chunks that are fed into the neural network. These samples can be accessed at Chen et al. (2025). Each chunk corresponds to a 0.23-s segment of audio. To simulate a global-level deviant, one chunk (the fifth #5) is deliberately altered, with the modified chunk highlighted in black for clarity. This violation is introduced by shuffling randomly but is constrained such that its rank representation is exclusive-i.e., not encountered during training-allowing it to function as a global-level violation.

These sound chunks are then transformed into rank-ordered representations as described previously (see Figs. 2 and 3). The rank sequences corresponding to each chunk are displayed in the top panel of Fig. 9b, with color coding that matches the chunk indices shown in Fig. 9a.

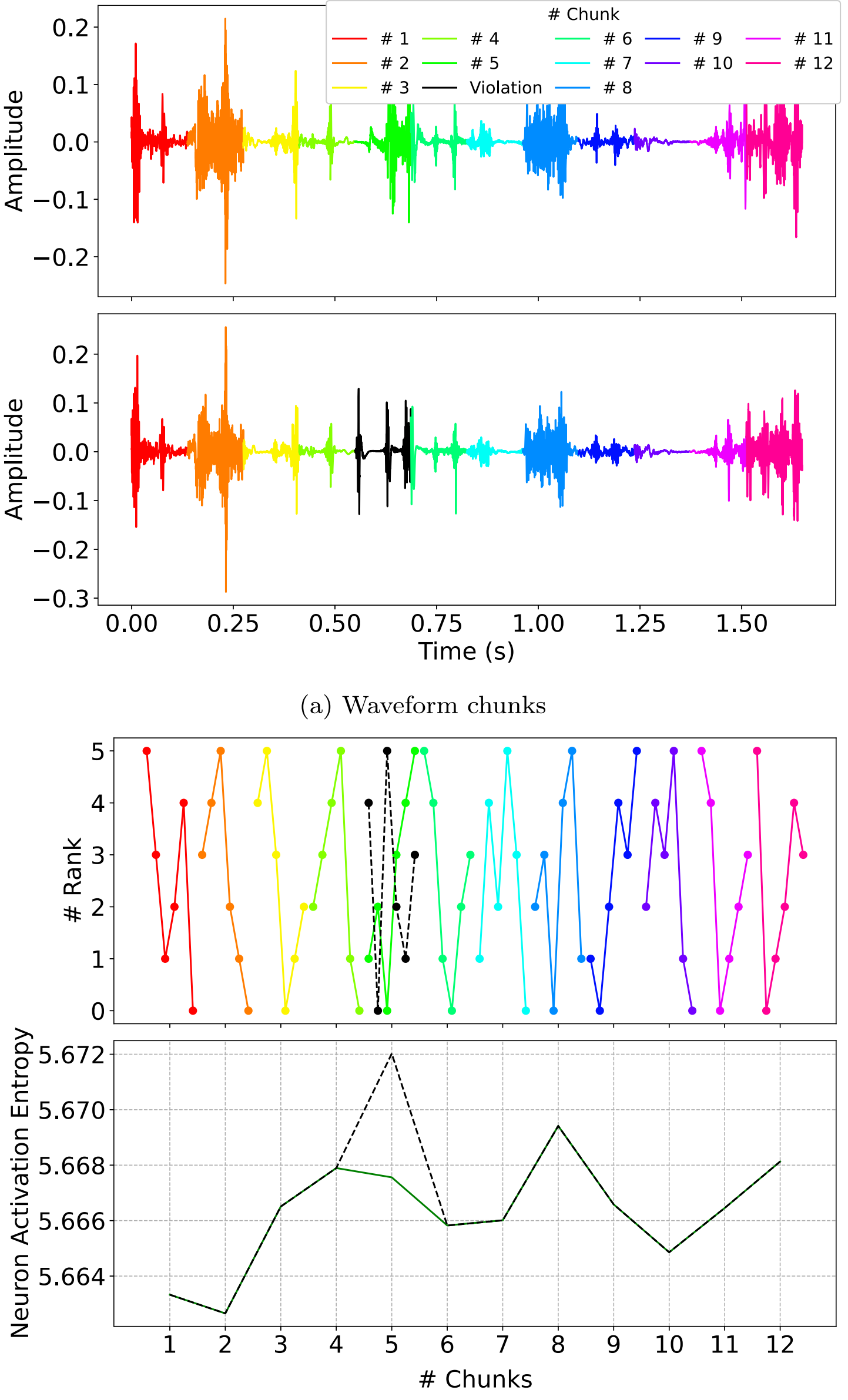

(b) Global response Dehaene et al. (2015) reproduction

**Fig. 9.** (a)Top: original sound sequence. Bottom: Global-level violation (drawn in black) replaced the #5 sound chunk. (b) Top: Rank pattern. Bottom: Entropy of neuron activities, reproduction of P3b-like wave in Dehaene et al. (2015).

Inspired by Shannon (1948), we compute the entropy of neuron activations in the rank layer, as entropy quantifies the uncertainty or disorder in a system, with higher entropy reflects a more diffuse or less confident activation pattern, which aligns closely with the neural processes underlying novelty detection: in cognitive neuroscience, the P3b component of event-related potentials is strongly associated with the conscious detection of unexpected, novel stimuli that violate higher-order expectations (Friedman et al., 2001). Thus, by measuring entropy, we aim to capture a computational correlate of the surprise or novelty-related signal reflected in P3b captured in Dehaene et al. (2015).

For comparison, we plot the entropy time course for sound streams with and without the introduced violation, as shown in the bottom panel of Fig. 9b. The black curve represents the entropy trajectory when the fifth chunk is replaced. As shown, a distinct entropy peak emerges at the fifth chunk position, indicating increased neural uncertainty or surprise-effectively reproducing the global novelty response observed in Dehaene et al. (2015) and highlighting the network's sensitivity to high-level sequence violations.

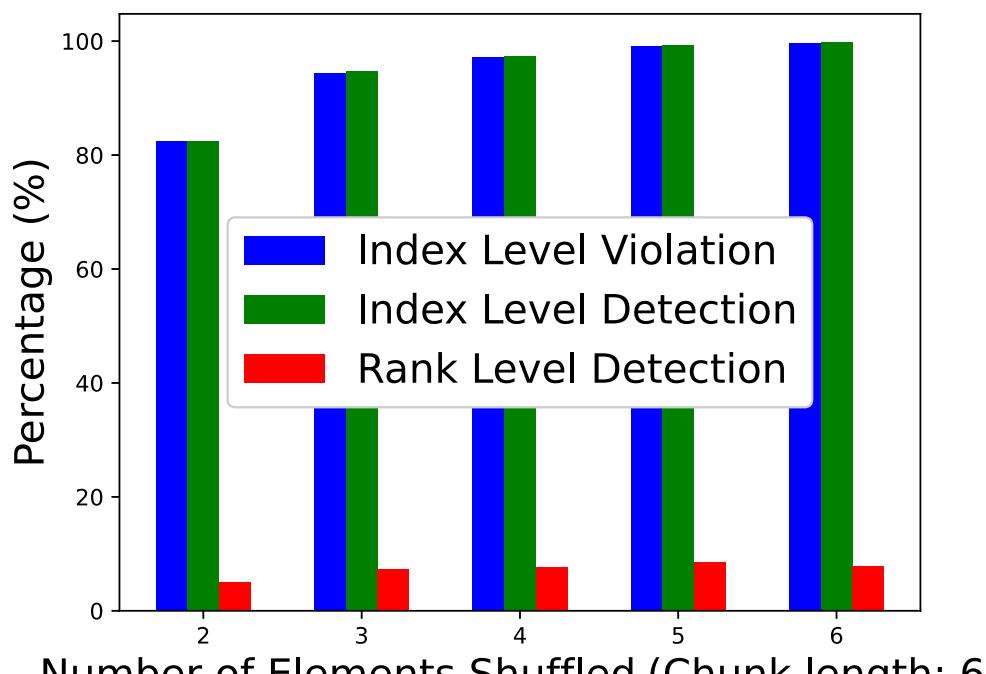

**Fig. 10.** Rank level generalization to item level violation.

**Table 1**
FP: False positives (e.g., flagged as violations when they should not be). FN: False negatives(e.g., violations that go undetected).

| Number of Swaps | Index FP (%) | Index FN (%) | Rank FP (%) | Rank FN (%) |
|---|---|---|---|---|
| 2 | 0.15 | 0.2 | 0.0 | 0.0 |
| 3 | 0.4 | 0.0 | 0.0 | 0.0 |
| 4 | 0.25 | 0.0 | 0.0 | 0.0 |
| 5 | 0.2 | 0.0 | 0.0 | 0.0 |
| 6 | 0.25 | 0.0 | 0.0 | 0.0 |

### *3.4. Robustness of rank-based representations to index-level perturbations*

The previous experiment demonstrated that our neural network can detect rank-level violations, replicating the cognitive "global response" effect. However, one may doubt whether this apparent generalization arises from insensitivity or underfitting, rather than from structured abstraction. To address this, we contrast rank-level generalization with the network's responsiveness to index-level novelty (e.g., unseen sequences).

The degree of index-level violation is controlled by the extent of derangement-that is, a completely new random permutation in which none of the n (where $n \in [2, 6]$) selected elements remain in their original positions-at the index level. Unlike the previous experiment that guaranteed all rank-level patterns were novel, here the degree of rank-level violation may vary with the extent of index shuffling. Each condition is tested on 2000 examples to ensure statistical reliability. Samples of each condition are available at Chen et al. (2025).

In Fig. 10, we plot the number of novel input sequences at the index level (i.e., sequences not seen during training) in blue, and the number of such novelties detected by the neural network via winner neuron changes, plotted in green. Both bars increase as more elements in the sequence are shuffled, showing strong sensitivity to index-level novelty. In contrast, the number of violations detected at the rank level (via decreases in winner neuron activation), plotted in red, remains relatively low and stable. This result suggests a form of rank-level robustness: the model is less sensitive at rank-level to permutations in the precise ordering of elements as long as their relative structure (i.e., rank pattern) is preserved.

To complement this analysis, we further evaluate the precision of our model's violation detection at both the index and rank levels, as summarized in Table 1. This evaluation includes False Positives (FP: cases incorrectly flagged as violations) and False Negatives (FN: cases that should have been flagged but were missed). The results show that detection at the rank level is remarkably stable, with zero observed errors across all conditions. Index-level detection, while slightly more error-prone, exhibits only minimal error rates ranging from 0% to 0.4%, indicating high precision overall. These results further highlight the clear distinction between index-level and rank-level violations observed in Fig. 10, reflecting the capacity of rank-order coding to support hierarchical generalization.

Together, these results implies that while the model encodes the raw symbol identity quite strictly (index-sensitive), it allows some tolerance or invariance at the rank level. This mimics aspects of cognitive processing, where structural patterns may be more resilient to surface-level variation.

## 4. Conclusion

In this work, we explored the use of rank-order coding as a mechanism for structure processing in neural networks as well as the hierarchical grammar role of rank. Section 3.2 demonstrates the network's ability to reconstruct complex temporal sequences from partial abstract input, revealing structure sensitivity-where rank-order coding encodes abstract, context-general rank representations that are dynamically instantiated as context-specific (Khanna et al., 2024) outputs during speech planning and ultimately realized as acoustic features during motor execution. This reflects the model's capacity to maintain generalizable representations while adapting them to specific temporal or segmental contexts. This behavior successfully mirrored the role of the pathway from LIFG (Broca's area) to premotor cortex in speech planning, as illustrated in the orange pathway in Fig. 1b. The reproduced "global response" (Dehaene et al., 2015) to structural violation in Section 3.3 supports the hypothesis that rank-based representation exhibits cognitive similarity to grammar that sensitive to hierarchical structure violation. Since the only alteration lies in the rank configuration, this response suggests that the rank-order chunk appears to function not merely as a compressed encoding, but as a medium encoding abstract. The experiment in Section 3.4 shows that the model demonstrates a clear distinction between index-level sensitivity and rank-level robustness, precisely encoding the identity and order of symbols while generalizing robustly to novel variations in their relative rank structure. This further confirms the hypothesis that the network tolerates index-level violation as long as the relative ordering of elements is preserved, implying that the rank layer abstracts away from raw input and encodes a kind of structural template or schema, similar to hierarchical generalization (Uddén et al., 2019). This mimics aspects of cognitive processing, where structural patterns may be more resilient to surface-level variation. This grammar-like behavior mirrors how human cognition detects and resists syntactic anomalies.

In future work, we will extend rank-order coding to model superordinate chunks, resembling nested tree structures. We also plan to investigate the cross-linguistic implications of rank-based encoding and explore its integration with muscle signals to further evaluate the capacity and generalization potential of rank-order coding.

## CRediT authorship contribution statement

**Xiaodan Chen:** Writing – review & editing, Writing – original draft, Visualization, Resources, Methodology, Investigation, Formal analysis, Data curation, Conceptualization; **Alexandre Pitti:** Writing – review & editing, Validation, Supervision, Project administration, Methodology, Investigation, Funding acquisition, Formal analysis, Conceptualization; **Mathias Quoy:** Writing – review & editing, Validation, Supervision, Project administration, Funding acquisition, Formal analysis, Conceptualization; **Nancy F. Chen:** Funding acquisition.

## Declaration of competing interest

Xiaodan Chen reports financial support was provided by ASTAR Research Entities. Mathias Quoy reports financial support was provided by National Centre for Scientific Research. If there are other authors, they declare that they have no known competing financial interests or personal relationships that could have appeared to influence the work reported in this paper.

## Acknowledgments

We would like to express our sincere gratitude to Krzysztof Lebioda who generously shared his source code. Xiaodan Chen is supported by CYU-IPAL and the Singapore A*STAR Research Attachment Programme. Mathias Quoy was supported by CNRS funding for a research semester at the IPAL Lab in Singapore. We also acknowledge the use of the Osaka compute-cluster resources of CY Cergy Paris University.